\documentclass[times, review, 10pt]{elsarticle}

\usepackage[numbers]{natbib}
\usepackage{algorithm}
\usepackage{algorithmic}

\usepackage{times}
\usepackage{helvet}
\usepackage{courier}

\usepackage{url}            
\usepackage{booktabs}       
\usepackage{amsfonts}       
\usepackage{nicefrac}       
\usepackage{microtype}      
\usepackage{xcolor}         
\usepackage{graphicx}
\usepackage{amsmath}
\usepackage{multirow}
\usepackage{subcaption}
\usepackage{pifont}

\def\tsc#1{\csdef{#1}{\textsc{\lowercase{#1}}\xspace}}
\tsc{WGM}
\tsc{QE}
\tsc{EP}
\tsc{PMS}
\tsc{BEC}
\tsc{DE}

\begin{document}
	\let\WriteBookmarks\relax
	\def\floatpagepagefraction{1}
	\def\textpagefraction{.001}

	
	\title {TimeFormer: Transformer with Attention Modulation Empowered by Temporal Characteristics for Time Series Forecasting}                      
	
	\address[1]{Software College, Northeastern University, Shenyang, China}
	\address[2]{School of Software, University of
    Science and Technology of China, Hefei, China}
    \address[3]{Xichang Satellite Launch Center, Xichang, China}
	
	\author[1]{Zhipeng Liu}
	\ead{2310543@stu.neu.edu.cn}
	
	\author[1]{Peibo Duan$^*$}
	\ead{duanpeibo@swc.neu.edu.cn}

    \author[1]{Xuan Tang}
	\ead{2471477@stu.neu.edu.cn}
    
    \author[1]{Baixin Li}
	\ead{libaixin@mails.neu.edu.cn}
    
    \author[1]{Yongsheng Huang}
	\ead{2371447@stu.neu.edu.cn}

    \author[3]{Mingyang Geng}
	\ead{gengmingyang13@nudt.edu.cn}
    
	\author[1]{Changsheng Zhang}
	\ead{zhangchangsheng@mail.neu.edu.cn}
	
	\author[1]{Bin Zhang}
	\ead{zhangbin@mail.neu.edu.cn}

    \author[2]{Binwu Wang$^*$}
	\ead{wbw2024@ustc.edu.cn}

	\cortext[cor1]{Corresponding author}

	\begin{abstract}
		Although Transformers excel in natural language processing, their extension to time series forecasting remains challenging due to insufficient consideration of the differences between textual and temporal modalities. In this paper, we develop a novel Transformer architecture designed for time series data, aiming to maximize its representational capacity. We identify two key but often overlooked characteristics of time series: (1) unidirectional influence from the past to the future, and (2) the phenomenon of decaying influence over time. These characteristics are introduced to enhance the attention mechanism of Transformers. We propose TimeFormer, whose core innovation is a self-attention mechanism with two modulation terms (MoSA), designed to capture these temporal priors of time series under the constraints of the Hawkes process and causal masking. Additionally, TimeFormer introduces a framework based on multi-scale and subsequence analysis to capture semantic dependencies at different temporal scales, enriching the temporal dependencies. Extensive experiments conducted on multiple real-world datasets show that TimeFormer significantly outperforms state-of-the-art methods, achieving up to a 7.45\% reduction in MSE compared to the best baseline and setting new benchmarks on 94.04\% of evaluation metrics. Moreover, we demonstrate that the MoSA mechanism can be broadly applied to enhance the performance of other Transformer-based models. 
		
	\end{abstract}

	\maketitle
	
	\section{Introduction}
    
Time series forecasting is of paramount importance, as it exerts a significant influence on various aspects of human life, such as energy distribution, quantitative investment and weather forecasting \cite{lim2021time,chen2023long,liu2025attributed}. However, the evolution of real-world systems is dynamic and complex, which poses a challenge for accurate prediction. Deep learning models, such as multi-layer perceptrons (MLPs) \cite{zhong2023multi}, recurrent neural networks (RNNs) \cite{kong2025unlocking}, convolutional neural networks (CNNs) \cite{wu2022timesnet}, and Transformers \cite{liu2023itransformer}, offer promising solutions for addressing these challenges.

In recent years, Transformers have become a cornerstone in deep learning due to their powerful representational capabilities, particularly through the self-attention mechanism that captures global dependencies in structured data. Initially successful in natural language processing (NLP), where they demonstrated exceptional performance and stability, Transformers have been adapted for time series forecasting through various enhanced architectures. Despite these efforts, Transformers have faced challenges in time series tasks \cite{tang2025unlocking}. Models like DLinear \cite{zeng2023transformers} and TimeMixer \cite{wang2024timemixer}, which rely on simple linear layers, have shown impressive results, prompting a reevaluation of Transformers in the time series community. This raises a critical question: why do Transformers excel in NLP but struggle with time series analysis?

\begin{figure*}[t]
    
    \begin{minipage}{0.48\textwidth}
        
        \includegraphics[width=\textwidth]{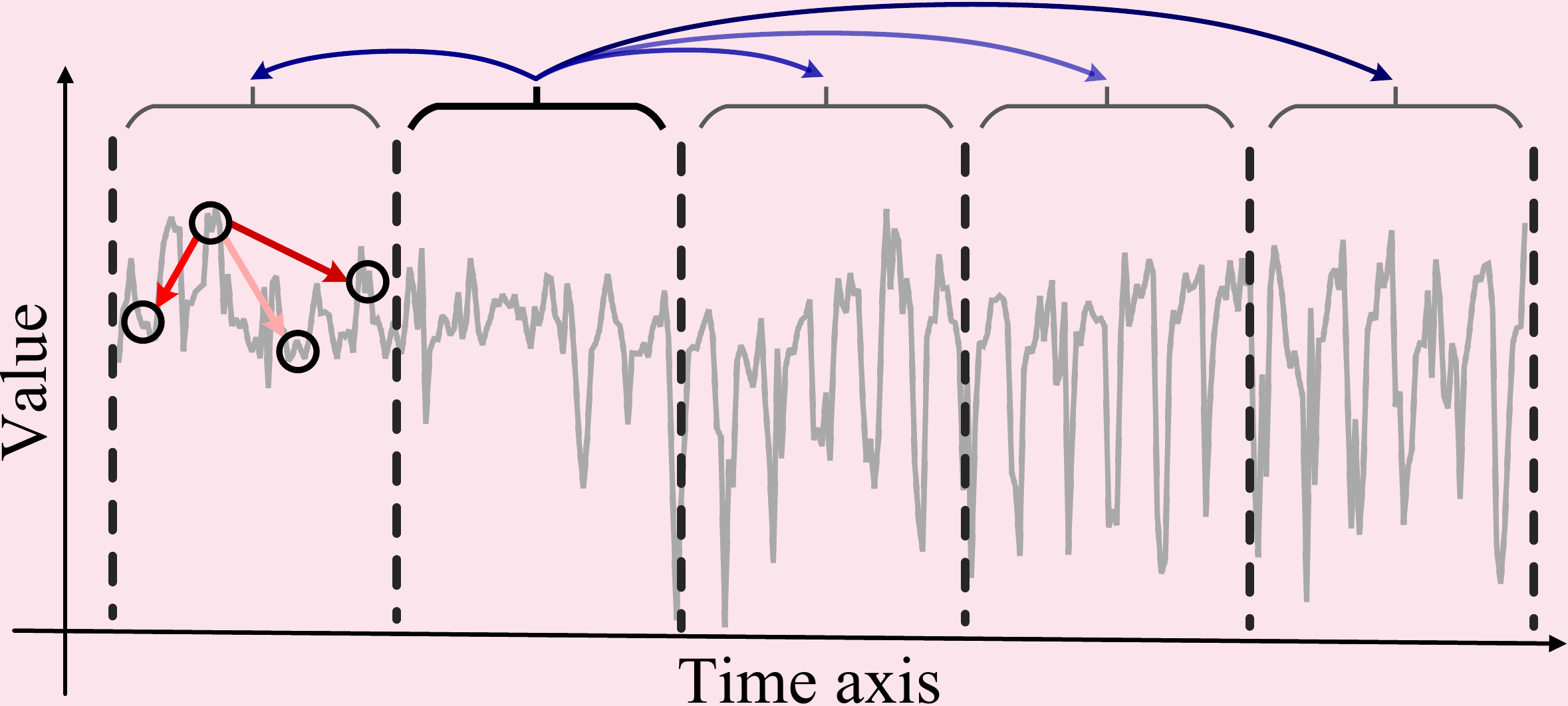}
        \subcaption{Conventional self-attention.}\label{fig:sub1}
    \end{minipage}\hspace{0.005\textwidth}
    \begin{minipage}{0.48\textwidth}
        
        \includegraphics[width=\textwidth]{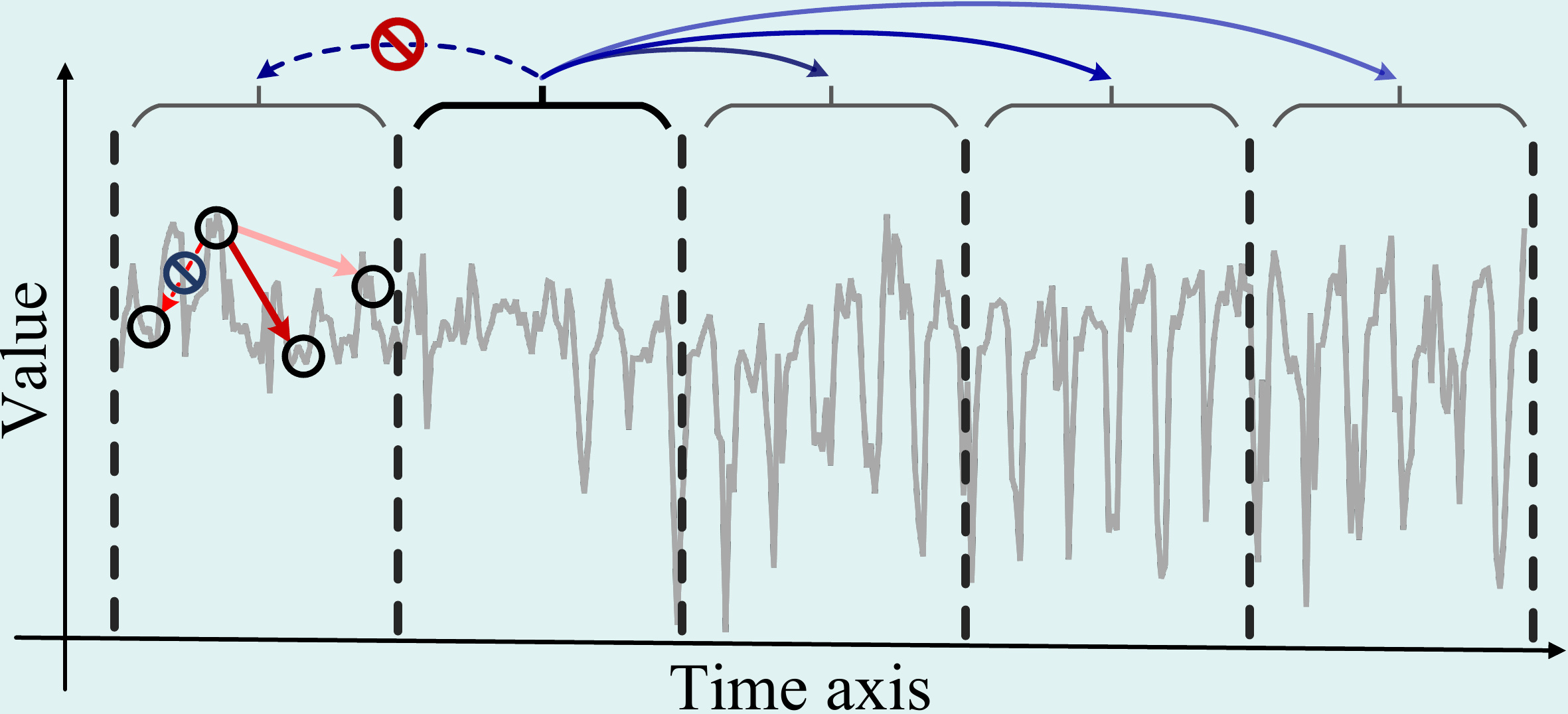}
        \subcaption{Self-attention with modulation terms (MoSA).}
        \label{fig:sub2}
    \end{minipage}
    \caption{Time series are divided into subsequences. Red and blue arrows indicate intra-subsequence and inter-subsequence dependencies, respectively, with color intensity reflecting the strength of the influence.}
    \label{intro}
\end{figure*}

We argue that existing Transformer-based forecasters fail to adequately account for the differences between time series data and natural language. Natural language contains rich word-level semantic information and contextual dependencies \cite{nie2022time}, crucial for the success of Transformers in natural language processing. For example, in “I like watermelon”, the words “I”, “like”, and “watermelon” function in a semantic manner as subject, verb, and object, respectively. In contrast, time series data comprises isolated numerical values with limited semantic meaning, and the dependencies between time steps are characterized by temporal properties rather than the contextual relationships in natural language. These differences pose challenges for modeling time series with standard Transformer architectures, since time series data inherently exhibit two temporal properties: (1) a unidirectional causal influence of the current time step on future steps, and (2) a decaying influence over subsequent time steps. Although existing studies attempt to capture semantic information within subsequences, the features extracted by MLPs \cite{nie2022time,tang2025unlocking} and conventional self-attention mechanisms \cite{chen2024pathformer} typically exhibit anti-temporal characteristics, as illustrated in Figure 1-(a). These issues motivate us to explore specialized mechanisms that equip self-attention with the ability to capture temporal properties (Figure 1-(b)).

In this paper, we propose TimeFormer, a simple yet effective Transformer architecture tailored to enhance Transformers’ representational capability for time series data by enriching semantic information and incorporating temporal properties. First, TimeFormer includes multi-scale temporal modeling, generating time series at different temporal resolutions from a hierarchical perspective. Subsequently, the time series at each scale is divided into subsequences to model temporal dependencies within and across these subsequences, respectively. This process is designed to enhance semantic representations and to reveal intrinsic semantic relationships, enabled by our novelly proposed self-attention mechanism, named MoSA. MoSA incorporates two modulation terms that align attention generation with temporal priors: (1) unidirectional causal dependencies and (2) decaying attention effects. Specifically, we design a temporal causal masking mechanism using a lower triangular mask matrix, preventing attention from future to past time steps, enforcing a unidirectional influence. Additionally, the Hawkes process is integrated into attention computation, penalizing distant time steps and emphasizing closer ones to simulate the decay effect. As illustrated in Figure 1, unlike conventional self-attention mechanisms that model bidirectional contextual dependency, MoSA explicitly incorporates these temporal priors, enabling more accurate modeling of temporal dynamics. Extensive experiments on real-world datasets demonstrate that our model achieves highly competitive forecasting performance. Furthermore, when the proposed MoSA mechanism is applied to other Transformer-based models, their performance improves, highlighting its broad applicability and potential. The contributions of this work can be summarized as follows:

\begin{itemize}
    \item We explore the potential for improving Transformer-based time series forecasting by analyzing the differences between natural language and time series data, leading to the development of TimeFormer, a simple yet effective architecture equipped with prior time series characteristics.

    \item TimeFormer employs a subsequence segmentation strategy and our proposed MoSA to enrich semantic representations and capture semantic dependencies. MoSA introduces a novel self-attention mechanism with two modulation terms, which incorporates causal masking to enforce unidirectional past-to-future dependencies, and adopts the Hawkes Process to model the decay of attention over time, thereby penalizing long-range dependencies.
    
    \item Extensive experiments on multiple real-world datasets show that TimeFormer achieves SOTA performance, setting new benchmarks on 94.04\% of evaluation metrics. Additionally, we integrate the proposed MoSA into other Transformer-based models, demonstrating its broad applicability and consistent performance gains.

\end{itemize}

The remainder of this paper is structured as follows. Section \ref{relatedwork} reviews the literature relevant to our approach. Section \ref{method} formulates the problem and details the proposed Timeformer framework. Section \ref{experiment} describes the experimental setup, including datasets, evaluation metrics, baseline methods, results, and further analyses. Finally, Finally, the conclusions are presented in Section \ref{conclusion}.

\begin{figure*}[!ht]
  \centering
  \includegraphics[scale=0.39]{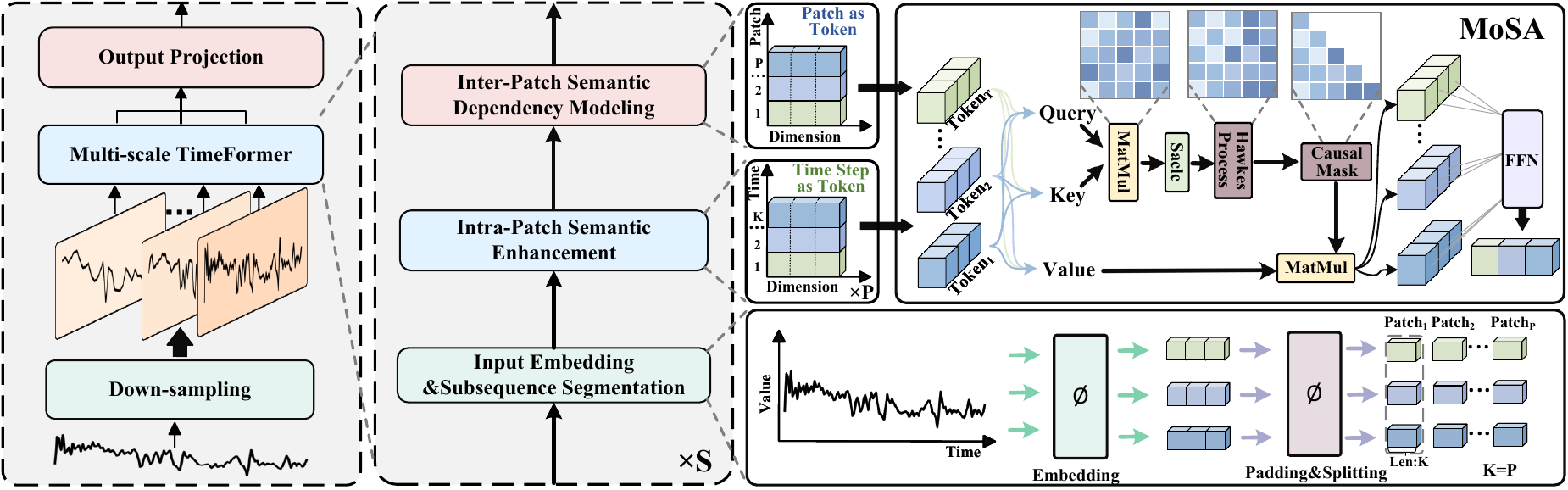}
  \caption{The framework of TimeFormer.}
  \label{fig2111}
\end{figure*}

\section{Related Works}
\label{relatedwork}

\subsection{Conventional Methods for Time Series Forecasting}

The task of multivariate time series forecasting has garnered significant attention from both industry and academia \cite{huang2023crossgnn,huang2024hdmixer,wang2024timexer,qiu2024tfb}. Statistical methods, such as Exponential Smoothing \cite{de2011forecasting} and ARIMA \cite{kalpakis2001distance}, offer efficient approaches for modeling temporal patterns. However, they are limited in capturing nonlinear characteristics. Among deep learning methods, CNN-based models leverage temporal convolutions to capture local temporal dependencies. For instance, TimesNet transforms the original one-dimensional time series into a two-dimensional representation and employs convolutions to capture multi-period features \cite{wu2022timesnet}. RNN models, such as LSTM and GRU, are specifically designed to capture temporal dependencies in sequential data \cite{liu2025disms}. For example, P-sLSTM introduces patching and channel independence to mitigate the short memory issue \cite{kong2025unlocking}. GNNs model spatial dependencies for prediction by analyzing the relationships between variables \cite{wang2024graph,wang2024fully}.  Emerging methods, such as LLM-based models, have also demonstrated effective performance in this field \cite{wang2024news}. Notably, recent MLP-based models exhibit superior prediction performance \cite{zeng2023transformers,Zhou2026TAMD,Wang2025TimeParticle,Nguyen2025ConEm,Li2025STNet}. For instance, Timemixer leverages the multi-scale nature of time series to separately integrate multi-scale trend and seasonal features \cite{wang2024timemixer}.

\subsection{Transformers for Time Series Forecasting}

Transformer \cite{vaswani2017attention,Yu2024Robformer,Luo2025TFDNet,Ye2026CVACL-MA,Xu2025FPF} has recently gained increasing attention in time series forecasting due to the similarity between natural language and time series, both of which are sequential in nature. Informer utilizes a self-attention mechanism with probabilistic selection of important time steps to reduce computational complexity \cite{zhou2021informer}. Autoformer introduces a novel decomposition block that separates trend and seasonal components, improving the modeling of long-range dependencies in time series data \cite{wu2021autoformer}. Fedformer combines the benefits of Fourier transform and attention mechanisms, addressing both global and local dependencies to enhance forecasting accuracy \cite{zhou2022fedformer}. Crossformer leverages cross-attention to model both global and local temporal dependencies in time series \cite{zhang2023crossformer}. iTransformer, inspired by GNN-based methods, treats variables as tokens and captures their dependencies to explore variable-level interactions \cite{liu2023itransformer}. CSformer employs a two-stage multi-headed self-attention mechanism to extract and integrate both channel-specific and sequence-specific information \cite{wang2025csformer}. Although they achieve significant progress in time series forecasting, they overlook the temporal characteristic of time series.

\section{Methodology}
\label{method}

\subsection{Problem Statement}
Let \( \mathbf{x}_t \in \mathbb{R}^N \) denote \( N\) variables of a regularly sampled time series at time $t$. Given the historical observations $\mathbf{X}=\{\mathbf{x}_t\}_{t=1}^{L_h} \in \mathbb{R}^{{L_h}\times N}$, the goal of multivariate time series forecasting task is to develop a model function for forecasting the future values $\mathbf{Y}=\{\mathbf{x}_t\}_{t={L_h}+1}^{{L_h}+{L_f}} \in \mathbb{R}^{{L_f}\times N}$, where ${L_h}$ and ${L_f}$ represent the lengths of the historical and future horizons, respectively. Mathematically, the problem can be formulated as follows:

\begin{equation}
    \mathbf{Y} = \mathcal{G}(\mathbf{X};\Theta),
\end{equation}
\noindent where $\mathcal{G}: \mathbb{R}^{L_h\times N}\rightarrow \mathbb{R}^{L_f\times N}$ represents the  model parameterized by $\Theta$.

\subsection{Overall Framework}
As shown in Figure \ref{fig2111}, the proposed TimeFormer adopts an encoder-only architecture based on the Transformer, and enhances representational capability through the principle of channel-independent learning \cite{nie2022time}. For simplicity, we introduce the framework using univariate setting (i.e., $\mathbf{X}\in\mathbb{R}^{L_h}$), which can be readily extended to multivariate cases by treating each variable independently. TimeFormer consists of three main components: \textit{Multi-Scale Sequence Sampling}, \textit{TimeFormer for Multi-Scale Temporal Modeling}, and \textit{Output Projection}. Mathematically, the process can be summarized as:
\begin{equation}
    \begin{aligned}
        \mathbf{X}^s &= {\text{Sampling}}_{(s)}(\mathbf{X}),\; \{s = 1,...,S\}\\
        \mathbf{Z}^{s} &= {\text{TimeFormer}}_{(s)}( \overline{\mathbf{X}}^{s}), \; \{s = 1,...,S\}\\
        \mathbf{\hat{Y}} &= {\text{Projection}}(\{\mathbf{Z}^s\}^S_{s=1}).
    \end{aligned}
\end{equation}

\noindent First, to capture more coarse-grained temporal patterns, such as trend information, the historical input series $\mathbf{X}$ is fed into $\text{Sampling}(\cdot)$ function, generating a set of $S$-scale time series via average pooling, denoted as $\mathcal{X}=\{\mathbf{X}^s\}_{s=1}^S$. Specifically, 
\begin{equation}
    \mathbf{X}^{s\in S}=\text{Avgpool}(\mathbf{X},\text{kernel}=s, \text{stride}=s),
\end{equation}
\noindent  where $ \mathbf{X}^{s\in S}\in \mathbb{R}^{\left\lceil\frac{L_h}{s}\right\rceil}$ represents the time series sampled at the $s$-th scale. Subsequently, we input the multi-scale time series into the proposed time formation module to capture temporal dependencies. The generated multi-scale time representation is denoted as $\mathcal{Z}=\{\mathbf{Z}^s\}^S_{s=1}\in\mathbb{R}^{S\times D_{model}}$. Finally, we use a fully connected neural network as the predictor, denoted as ${\text{Projection}}(\cdot)$, to decode $\text{Flatten}( \mathcal{Z})$ and predict future time series. The core of this model is the proposed multi-scale analysis-based TimeFormer, which we will discuss in detail in the next section.

\subsection{TimeFormer for Multi-scale Temporal Modeling}

The TimeFormer module processes each specific scale sequence ${\mathbf{X}}^{s \in S} \in \mathbb{R}^{\left\lceil \frac{L_h}{s} \right\rceil}$ independently and consists of four main components: the input embedding module, the subsequence segmentation module, the semantic enhancement module, and the semantic dependency modeling module. The latter two models are implemented using our proposed \underline{S}elf-\underline{A}ttention mechanism with \underline{Mo}dulation term (MoSA). To simplify notation for convenience in explanation, we use the $s$-th scale as an example and denote $\left\lceil \frac{L_h}{s} \right\rceil$ as $L_s$.

\subsubsection{Input Embedding}

We employ a 1D-CNN module, which is denoted as ${\text{Embedding}}_{(s)}(\cdot): \mathbb{R}^{L_s} \rightarrow \mathbb{R}^{{L_s} \times D_{model}}$, to capture high-dimensional representations of the variable while preserving their temporal characteristics, generating $\overline{\mathbf{X}}^s\in \mathbb{R}^{L_s\times D_{model}}$,

\begin{equation}
    \overline{\mathbf{X}}^s = {\text{Embedding}}_{(s)}(\mathbf{X}^s).
\end{equation}

\subsubsection{Subsequence Segmentation}

Inspired by the patching method from \cite{nie2022time}, we split $\overline{\mathbf{X}}^s$ into $P_s$ non-overlapping subsequences, each containing $K_s$ time steps, as using overlapping subsequences might lead to redundancy in temporal features. Additionally, we employ the proposed MoSA to capture temporal dependencies within and between subsequences. The model’s performance is sensitive to the length of the patches: too many tokens can lead to diminished early weights, while too few may hinder the model’s ability to capture rich temporal patterns. Thus, we strike a balance between $P_s$ and $K_s$ by setting them to be equal, i.e., $P_s = K_s = \left\lceil \sqrt{L_s} \right\rceil$. To achieve this, we apply zero-padding to $\overline{\mathbf{X}}^s$, extending the time steps from $L_s$ to $P_s \times K_s$. The set of subsequences, $\mathcal{P}^s=\{{\mathbf{P}_p^s\}}_{p=1}^{P_s}\in \mathbb{R}^{P_s\times K_s\times D_{model}}$, can be obtained as:
\begin{equation}  \mathbf{P}_1^s,\mathbf{P}_2^s,...,\mathbf{P}_{P}^s=\text{Splitting}(\text{Padding}(\overline{\mathbf{X}}^s)).
\end{equation}

\subsubsection{Intra-Patch Semantic Enhancement}
\label{intra}
To address the issues of flawed temporal modeling of subsequences learned by linear models or the conventional self-attention mechanism \cite{nie2022time,chen2024pathformer}, we employ the proposed MoSA, which would be introduced in the next section, to capture temporal dependencies of each subsequence in $\mathcal{P}^s$, which treats the features at each time step as tokens to enhance semantic representation, generating $\widehat{\mathbf{H}}^s_{p\in P}\in \mathbb{R}^{K_s\times D_{model}}$ as follows,
\begin{equation}
    \widehat{\mathbf{H}}^s_{p}=\text{MoSA}_{(intra)}^s(\mathbf{P}^s_p).
\end{equation}
Subsequently, to facilitate the subsequent modeling of inter-subsequence temporal dependencies, we employ a feedforward neural network (FNN) to project $\widehat{\mathbf{H}}_{p}^s$ into a vector representation $\widetilde{\mathbf{H}}^s_p\in\mathbb{R}^{D_{model}}$,
\begin{equation}
    {\widetilde{\mathbf{H}}}^s_p=\text{FNN}(\text{Flatten}({\widehat{\mathbf{H}}}^s_{p})).
\end{equation}
\noindent $\widetilde{\mathbf{H}}^s=\{ \widetilde{\mathbf{H}}^s_p\}_{p=1}^{P_s}\in\mathbb{R}^{P_s\times D_{model}}$ denotes the set of temporal representations for all subsequences.

\subsubsection{Inter-Patch Semantic Dependency Modeling}
\label{inter}
With the rich semantic information of each subsequence, we aim to further capture global temporal dependencies by analyzing the semantic dependencies among subsequences. This can also be achieved by employing MoSA, which treats each subsequence as a token to model temporal dependencies, generating $\mathbf{H}^s\in \mathbb{R}^{P_s\times D_{model}}$,

\begin{equation}
    {\mathbf{H}^s}=\text{MoSA}_{(inter)}^s(\widetilde{\mathbf{H}}^s).
\end{equation}

\noindent Last, to ensure consistent dimensionality across all scales for output projection, we use a feedforward neural network to generate the final temporal representation $\mathbf{Z}\in\mathbb{R}^{D_{model}}$,
\begin{equation}
    \mathbf{Z}^s=\text{FFN}(\text{Flatten}(\mathbf{H}^s)).
\end{equation}

\subsection{Self-Attention with Modulation Term (MoSA)}
We propose a novel self-attention mechanism for time series modeling, called MoSA, which incorporates two temporal characteristics: \textbf{attention decay} and \textbf{unidirectional past-to-future influence}. These enhancements facilitate better modeling of temporal dependencies in time series data. Given a time series $\widehat{\mathbf{X}}\in\mathbb{R}^{T\times F}=\{\hat{\mathbf{x}}_{t}\}_{t=1}^T$, where $T$ and $F$ represent the sequence length and feature dimension, each token $\hat{\mathbf{x}}_{t\in T}$ is projected into the query, key, and value spaces through linear transformations:

\begin{equation}
\mathbf{q}_t = \hat{\mathbf{x}}_{t} {W}_q, \quad \mathbf{k}_t = \hat{\mathbf{x}}_{t} {W}_k, \quad \mathbf{v}_t = \hat{\mathbf{x}}_{t} {W}_v,
\end{equation}

\noindent where ${W}_q, {W}_k, \in \mathbb{R}^{F \times d}, {W}_v \in \mathbb{R}^{F \times d_{m}}$ are learnable parameters, $d$ is the dimension of $\mathbf{q}_t$ and $\mathbf{k}_t$, and $d_{m}$ is the dimension of $\mathbf{v}_t$. The pairwise correlation between token $i$ and $j$ is obtained using the scaled dot-product of their query and key vectors:

\begin{equation}
\mathbf{A}_{i,j} = \frac{\exp(\frac{\mathbf{q}_i \cdot \mathbf{k}_j^\top}{\sqrt{d}})}{\sum_{\tau=1}^{T} \exp(\frac{\mathbf{q}_i \cdot \mathbf{k}_{\tau}^\top}{\sqrt{d}})},
\end{equation}
\noindent where $\mathbf{A}_{i,j}$ is the element of the attention matrix $\mathbf{A}\in \mathbb{R}^{T\times T}$, and $d$ is the dimension of the key vector used for scaling. 

\subsubsection{Attention Decay with Hawkes Process} Temporal influences frequently exhibit a decaying property, meaning the impact of specific events tends to diminish over time. To incorporate this characteristic, we introduce the Hawkes process into the self-attention mechanism, which encourages the model to produce decaying attention weights, imposing sparse effects on the attention scores between distant query-key pairs. Specifically, $\widehat{\mathbf{A}}_{i,j}$ can be reformulated as:

\begin{equation}
\widehat{\mathbf{A}}_{i,j} = \frac{\exp(\frac{\mathbf{q}_i \cdot \mathbf{k}_j^\top}{\sqrt{d}})}{\sum_{\tau=1}^{T} \exp(\frac{\mathbf{q}_i \cdot \mathbf{k}_{\tau}^\top}{\sqrt{d}})}\times \Omega_{i,j},
\end{equation}
where $\Omega_{i,j}$ represents the modulation term generated by Hawkes process. 

\noindent\textbf{Theory 1. Hawkes Process} The Hawkes process is a mathematical model used to characterize self-exciting processes. It models a sequence of events where the occurrence of each event increases (or excites) the likelihood of subsequent events, with this excitatory effect decaying over time. Given the time steps at which the observed sequence occurs $\{t_1, t_2, \cdots\}$, the intensity function of the Hawkes process in discrete form is:

\begin{equation}
\Omega_{i,j}=\mu+\sum \epsilon e^{-\gamma\left(t_j-t_i\right)},
\end{equation}

\noindent where \(\mu\) and \(\epsilon\) represent the background intensity and impact factor. $\gamma$ is the decay rate of the excitatory effect over time. The term $\Omega_{i,j}$ models the intensity at time step $t_j$ caused by events occurring at previous time steps $t_i$, with the summation accounting for the influence of all prior events. It is noted that, as this study aims to explore temporal dependencies, we assume no external potential interference, thereby setting $\mu = 0$. Additionally, since the attention matrix inherently captures the influence of historical events, $\epsilon = 1$ imposes no need for additional weighting.

\subsubsection{Unidirectional Past-to-future Influence}

In the traditional Transformer, the self-attention mechanism captures bidirectional interactions between token pairs. However, temporal data typically follows a unidirectional dependency from past to future. This fundamental difference makes the standard self-attention mechanism unsuitable for temporal modeling, as it allows access to future information and violates the general principle of temporal causality. To address this issue, we apply a causal (i.e., lower triangular) mask to $\widehat{\mathbf{A}}$, restricting each token to attend only to itself and previous tokens. The final correlation $\widetilde{\mathbf{A}}_{i,j} \in \mathbf{A}$ between token $i$ and $j$ is represented as follows:

\begin{equation} 
\label{aggregation}
\widetilde{\mathbf{A}}_{i,j} = \text{Mask}(\widehat{\mathbf{A}}_{i,j})=\begin{cases} \widehat{\mathbf{A}}_{i,j}, & j \leq i \ \\
0, & j > i \end{cases}. 
\end{equation}

In conclusion, the temporal self-attention mechanism can be denoted as $\sum_{\tau=1}^{T} \mathbf{A}_{t,\tau} \cdot \mathbf{v}_\tau$.

Notably, in contrast to the vanilla self-attention in Transformers, MoSA captures global dependencies while incorporating temporal characteristics. Like the vanilla Transformer, MoSA also integrates multi-head attention, BatchNorm, and residual connections.

\section{Experiment}

\label{experiment}

\subsection{Experimental Setup}
\subsubsection{Datasets}

To evaluate the effectiveness of the proposed TimeFormer, we conduct extensive experiments on seven multivariate time series datasets, namely ETTh1, ETTh2, ETTm1, ETTm2, Exchange, Weather, and Electricity, which serve as standard benchmarks in previous works \cite{chen2024pathformer,Tan2025series}.

\begin{table}[t]
    \centering
    \resizebox{0.8\linewidth}{!}{%
    \begin{tabular}{lccc}
        \toprule
        {Dataset} & {Dataset Size} & {Dimension} & {Sampling Rate} \\
        \midrule
        ETTm1       & (34465, 11521, 11521) & 7   & 15 min \\
        ETTm2       & (34465, 11521, 11521) & 7   & 15 min \\
        ETTh1       & (8545, 2881, 2881)    & 7   & 1 h \\
        ETTh2       & (8545, 2881, 2881)    & 7   & 1 h \\
        Exchange   & (5120, 665, 1422)     & 8   & 1 day \\
        Weather    & (36792, 5271, 10540)  & 21  & 10 min \\
        Electricity& (18317, 2633, 5261)   & 321 & 1 h \\
        \bottomrule
    \end{tabular}%
    }
    \caption{Statistics of datasets.}
    \label{dataset}
\end{table}

Specifically, ETTh and ETTm are derived from the Electricity Transformer Temperature (ETT) dataset, which records transformer load and ambient temperature variations at hourly and minute resolutions, reflecting the influence of periodic load patterns and environmental conditions on energy systems.\footnote{https://github.com/zhouhaoyi/ETDataset.} The Exchange dataset consists of exchange rate fluctuations among multiple currencies, exhibiting strong non-stationarity and stochastic volatility.\footnote{https://github.com/laiguokun/multivariate-time-series-data.} The Weather dataset contains multiple meteorological variables (such as temperature, humidity, and wind speed), allowing the evaluation of forecasting performance on complex natural systems.\footnote{https://www.bgc-jena.mpg.de/wetter.} The Electricity dataset records power consumption from hundreds of clients, representing a high-dimensional, strongly periodic, and noisy industrial dataset.\footnote{https://archive.ics.uci.edu/ml/datasets/ElectricityLoadDiagrams20112014.} Table \ref{dataset} summarizes the details of the datasets.

\subsubsection{Evaluation Metrics}
\label{sec:evaluation_metrics}

Following prior studies \cite{Beltran2025series,Peng2026ConEm,Zhan2024series}, to evaluate the forecasting performance of {TimeFormer}, we employ two widely used metrics: {Mean Squared Error (MSE)} and {Mean Absolute Error (MAE)}. 
Given ground-truth future sequences $\mathbf{Y} \in \mathbb{R}^{L_f \times N}$ and predictions $\widehat{\mathbf{Y}} \in \mathbb{R}^{L_f \times N}$, where $L_f$ and $N$ denote the prediction length and variable dimension respectively, the two metrics are defined as:

\begin{equation}
\mathrm{MSE} 
= \frac{1}{L_f N}
\sum_{t=1}^{L_f}\sum_{n=1}^{N}
\big(\widehat{Y}_{t,n}-Y_{t,n}\big)^2,
\label{eq:mse}
\end{equation}

\begin{equation}
\mathrm{MAE} 
= \frac{1}{L_f N}
\sum_{t=1}^{L_f}\sum_{n=1}^{N}
\big|\widehat{Y}_{t,n}-Y_{t,n}\big|.
\label{eq:mae}
\end{equation}

\noindent where MSE emphasizes large deviations due to its quadratic form, while MAE offers robustness against outliers. 
All experiments are repeated five times, and the averaged MSE and MAE are reported to ensure fair and stable comparison.

\subsubsection{Baselines}

We compare our model with three advanced \textbf{MLP-based predictors}: \textbf{Timemixer++} \cite{wang2024timemixer++}, \textbf{PatchMLP} \cite{tang2025unlocking}, \textbf{TimeMixer} \cite{wang2024timemixer}, and \textbf{DLinear} \cite{zeng2023transformers}, which demonstrate impressive performance in recent time series forecasting studies. We also include five widely adopted \textbf{Transformer-based methods}: \textbf{iTransformer} \cite{liu2023itransformer}, \textbf{PatchTST} \cite{nie2022time}, \textbf{Crossformer} \cite{zhang2023crossformer}, \textbf{FEDformer} \cite{zhou2022fedformer}, and \textbf{Autoformer} \cite{wu2021autoformer}.

For MLP-based models, \textbf{DLinear} uses channel-wise linear projections for simple yet effective forecasting, \textbf{TimeMixer} integrates temporal dependencies through hierarchical mixing layers, \textbf{PatchMLP} employs patch-wise embeddings for efficient spatio-temporal modeling, and \textbf{Timemixer++} enhances TimeMixer with multi-resolution mixing for universal time-series modeling.
This comprehensive set of baselines enables fair evaluation across diverse modeling paradigms.

The Transformer-based baselines exploit various attention mechanisms for pair-wise dependency modeling. 
\textbf{Autoformer} adopts series decomposition to capture trend and seasonal components, \textbf{FEDformer} performs frequency-domain attention for efficiency, while \textbf{Crossformer} models cross-dimensional interactions. 
\textbf{PatchTST} applies patch-level tokenization for local temporal learning, and \textbf{iTransformer} treats each variable as an independent token to enhance multivariate representation.

\subsubsection{Implementation Details}
We adopt the same data preprocessing approach, including the train/validation/test split ratio, as used in \cite{wu2021autoformer, wu2022timesnet}, to ensure a fair comparison. Experiments are conducted with different prediction lengths \( L_f \in \{24, 48, 96, 192, 336, 720\} \), and the look-back window is set to $L_h$ = 96. All experiments are executed on a single NVIDIA GeForce RTX 4090 GPU using PyCharm 3.8 and PyTorch 2.1. The Adam optimizer \cite{kingma2014adam} is utilized with an initial learning rate of 0.005, and models are trained for 100 epochs. Following previous studies, we utilize Mean Squared Error (MSE) and Mean Absolute Error (MAE) as metrics to assess the performance. And each experiment is repeated five times independently, with average MSE and MAE reported for comparison.

\begin{table}[t]
\renewcommand{\arraystretch}{0.8}
\footnotesize	
\centering
\setlength{\tabcolsep}{3pt}
\resizebox{\textwidth}{!}{
\begin{tabular}
{c|c|cc|cc|cc|cc|cc|cc|cc|cc|cc|cc}
\toprule[1.pt]

\multicolumn{2}{c}{Methods}& \multicolumn{2}{c}{Ours}& \multicolumn{2}{c}{Timemixer++}& \multicolumn{2}{c}{PatchMLP}&\multicolumn{2}{c}{TimeMixer}& \multicolumn{2}{c}{iTransformer}& \multicolumn{2}{c}{PatchTST}&  \multicolumn{2}{c}{Crossformer}& \multicolumn{2}{c}{DLinear}& \multicolumn{2}{c}{FEDformer}& \multicolumn{2}{c}{Autoformer}\\ 
\cmidrule(lr){1-2}\cmidrule(lr){3-4}\cmidrule(lr){5-6}\cmidrule(lr){7-8}\cmidrule(lr){9-10}\cmidrule(lr){11-12}\cmidrule(lr){13-14}\cmidrule(lr){15-16}\cmidrule(lr){17-18}\cmidrule(lr){19-20}\cmidrule(lr){21-22}
\multicolumn{2}{c}{Metric}& MSE&MAE& MSE&MAE& MSE&MAE& MSE&MAE& MSE&MAE& MSE&MAE& MSE&MAE& MSE&MAE& MSE&MAE& MSE&MAE\\
\toprule[1.pt]
\multirow{7}{*}{\rotatebox{90}{ETTm1}}
&24&\textbf{0.204}&\textbf{0.281}&\underline{0.209}&\underline{0.285}&0.215&0.289&0.215&0.290&0.224&0.290&0.213&0.287&0.244&0.316&0.257&0.326&0.382&0.419&0.425&0.434\\
&48&\textbf{0.275}&\textbf{0.329}&\underline{0.287}&\underline{0.337}&0.289&0.339&\underline{0.287}&0.338&0.299&0.343&0.292&0.340&0.311&0.355&0.317&0.357&0.467&0.454&0.494&0.477\\
&96&\textbf{0.311}&\textbf{0.354}&0.325&0.365&\underline{0.324}&\underline{0.363}&0.328&0.366&0.331&0.366&0.328&0.366&0.431&0.431&0.344&0.373&0.434&0.447&0.597&0.503\\
&192&\textbf{0.349}&\textbf{0.377}&\underline{0.363}&\underline{0.377}&0.368&0.387&0.366&0.385&0.384&0.393&0.366&0.386&0.492&0.460&0.387&0.397&0.486&0.486&0.631&0.521\\
&336&\textbf{0.377}&\textbf{0.399}&\underline{0.386}&0.411&0.405&0.415&0.397&0.407&0.438&0.426&0.392&\underline{0.405}&0.657&0.576&0.418&0.421&0.470&0.480&0.541&0.515\\
&720&\textbf{0.439}&\textbf{0.436}&0.459&0.442&0.477&0.446&0.458&0.443&0.497&0.460&\underline{0.452}&\underline{0.440}&0.756&0.663&0.478&0.458&0.561&0.516&0.703&0.578\\
\cmidrule(lr){2-22}
&Avg&\textbf{0.325}&\textbf{0.362}&\underline{0.338}&\underline{0.369}&0.346&0.373&0.341&0.371&0.362&0.379&0.340&0.370&0.481&0.466&0.366&0.388&0.466&0.467&0.565&0.504\\

\midrule[0.7pt]

\multirow{7}{*}{\rotatebox{90}{ETTm2}}
&24&\textbf{0.092}&\textbf{0.173}&0.104&0.203&\underline{0.102}&\underline{0.198}&0.116&0.217&0.119&0.220&0.118&0.219&0.138&0.244&0.112&0.216&0.148&0.252&0.150&0.250\\
&48&\textbf{0.133}&\textbf{0.226}&0.151&0.248&\underline{0.143}&\underline{0.234}&0.156&0.253&0.156&0.253&0.159&0.256&0.198&0.300&0.158&0.264&0.172&0.267&0.173&0.268\\
&96 &\textbf{0.165}&\textbf{0.249}&0.175&\underline{0.255}&0.182&0.264&\underline{0.174}&0.256&0.180&0.264&0.178&0.262&0.324&0.374&0.185&0.276&0.205&0.287&0.211&0.291\\
&192&\textbf{0.220}&\textbf{0.276}&0.245&\underline{0.298}&0.243&0.308&\underline{0.239}&0.300&0.249&0.310&0.245&0.307&0.376&0.415&0.253&0.322&0.268&0.326&0.269&0.326\\
&336&\textbf{0.277}&\textbf{0.326}&0.296&\underline{0.332}&0.313&0.351&\underline{0.295}&0.338&0.315&0.353&0.309&0.348&0.575&0.552&0.322&0.373&0.329&0.365&0.328&0.364\\
&720&\textbf{0.382}&\textbf{0.384}&0.440&0.418&0.414&0.407&\underline{0.393}&\underline{0.395}&0.414&0.408&0.409&0.405&0.735&0.635&0.468&0.462&0.428&0.424&0.428&0.422\\
\cmidrule(lr){2-22}
&Avg&\textbf{0.211}&\textbf{0.272}&0.235&\underline{0.292}&0.232&0.293&\underline{0.228}&0.293&0.238&0.301&0.236&0.299&0.391&0.420&0.249&0.318&0.258&0.320&0.259&0.320\\
\midrule[0.7pt]

\multirow{7}{*}{\rotatebox{90}{ETTh1}}
&24&\textbf{0.292}&\textbf{0.349}&0.310&0.360&0.309&\underline{0.358}&{0.312}&0.361&0.317&0.363&\underline{0.307}&0.359&0.326&0.376&0.329&0.374&0.392&0.433&0.362&0.411\\
&48&\textbf{0.335}&\textbf{0.374}&0.347&0.384&0.350&0.385&\underline{0.344}&\underline{0.383}&0.358&0.388&0.353&0.388&0.366&0.405&0.362&0.394&0.420&0.448&0.454&0.461\\
& 96&\textbf{0.374}&\textbf{0.398}&0.397&\textbf{0.398}&0.393&0.405&0.388&0.405&\underline{0.385}&\underline{0.404}&0.389&0.409&0.413&0.445&0.395&0.410&0.415&0.448&0.472&0.469\\
&192&\textbf{0.431}&0.434&0.451&\textbf{0.429}&0.443&\underline{0.432}&0.456&0.444&\underline{0.436}&0.434&0.442&0.440&0.553&0.535&0.448&0.440&0.550&0.515&0.526&0.495\\
&336&\textbf{0.468}&\textbf{0.457}&0.494&0.467&0.498&0.464&0.481&0.464&\underline{0.480}&\underline{0.459}&0.495&0.469&0.717&0.656&0.509&0.485&0.492&0.492&0.524&0.500\\
&720&\textbf{0.464}&\textbf{0.470}&0.498&\underline{0.476}&0.519&0.494&0.504&0.485&\underline{0.496}&0.480&0.502&0.489&0.762&0.693&0.548&0.532&0.505&0.512&0.530&0.524\\
\cmidrule(lr){2-22}
&Avg&\textbf{0.394}&\textbf{0.413}&0.416&\underline{0.419}&0.418&0.422&0.414&0.423&\underline{0.412}&0.421&0.414&0.425&0.522&0.518&0.431&0.439&0.462&0.474&0.478&0.476\\
\midrule[0.7pt]

\multirow{7}{*}{\rotatebox{90}{ETTh2}}
&24&\textbf{0.174}&\textbf{0.265}&\underline{0.178}&\underline{0.267}&0.182&0.273&0.188&0.277&0.191&0.284&0.187&0.277&0.305&0.382&0.186&0.282&0.260&0.340&0.262&0.339\\
&48&\textbf{0.229}&\textbf{0.307}&\underline{0.232}&\textbf{0.307}&0.234&\underline{0.308}&0.240&0.311&0.260&0.332&0.242&0.312&0.376&0.427&0.254&0.332&0.292&0.360&0.294&0.359\\
&96 &\textbf{0.288}&\textbf{0.336}&0.295&\underline{0.338}&0.301&0.352&\underline{0.294}&0.346&0.297&0.348&0.303&0.352&0.613&0.536&0.313&0.366&0.342&0.387&0.350&0.392\\
&192&\textbf{0.367}&\textbf{0.382}&\underline{0.371}&\underline{0.386}&0.382&0.400&0.378&0.398&0.377&0.398&0.379&0.400&0.653&0.558&0.424&0.435&0.427&0.438&0.426&0.437\\
&336&\textbf{0.417}&\textbf{0.425}&0.422&\underline{0.428}&0.437&0.440&\underline{0.421}&0.433&0.424&0.432&0.428&0.440&0.725&0.708&0.454&0.460&0.470&0.478&0.461&0.473\\
&720&\textbf{0.425}&\textbf{0.438}&\underline{0.430}&\underline{0.441}&0.451&0.458&0.455&0.463&0.432&0.446&0.435&0.451&0.962&0.735&0.580&0.539&0.484&0.489&0.468&0.478\\
\cmidrule(lr){2-22}
&Avg&\textbf{0.316}&\textbf{0.358}&\underline{0.321}&\underline{0.361}&0.331&0.372&0.329&0.371&0.330&0.373&0.392&0.372&0.605&0.557&0.368&0.402&0.379&0.415&0.376&0.413\\
\midrule[0.7pt]

\multirow{7}{*}{\rotatebox{90}{Exchange}}
&24&\textbf{0.023}&\textbf{0.103}&0.029&0.116&0.026&0.111&\underline{0.025}&\underline{0.109}&0.026&0.111&0.065&0.181&0.082&0.199&0.026&0.111&0.077&0.205&0.099&0.230\\
&48&\textbf{0.042}&\textbf{0.144}&0.047&0.150&0.048&0.152&\underline{0.045}&\underline{0.145}&0.046&0.148&0.100&0.228&0.113&0.239&0.049&0.157&0.134&0.258&0.114&0.246\\
&96 &\textbf{0.084}&\textbf{0.204}&\underline{0.086}&0.206&0.092&0.213&\underline{0.086}&\underline{0.205}&0.096&0.219&0.089&0.207&0.294&0.402&0.097&0.221&0.183&0.311&0.175&0.302\\
&192&\textbf{0.180}&\textbf{0.301}&0.184&0.307&\textbf{0.180}&0.304&0.183&\underline{0.303}&0.183&0.305&0.184&0.305&0.519&0.561&0.241&0.354&0.321&0.411&0.275&0.379\\
&336&0.342&0.423&0.351&0.432&0.352&0.429&0.367&0.439&\textbf{0.333}&\textbf{0.418}&\underline{0.339}&\underline{0.421}&0.693&0.654&0.344&0.437&0.428&0.485&0.468&0.899\\
&720&\textbf{0.888}&\textbf{0.708}&0.901&0.714&\underline{0.895}&\underline{0.713}&0.899&0.714&0.897&0.718&0.908&0.718&1.625&1.039&0.972&0.781&1.338&0.899&1.133&0.845\\
\cmidrule(lr){2-22}
&Avg&\textbf{0.259}&\textbf{0.313}&0.266&0.320&0.265&0.320&0.267&0.319&\underline{0.263}&\underline{0.319}&0.280&0.343&0.387&0.515&0.288&0.343&0.413&0.428&0.377&0.483\\

\midrule[0.7pt]

\multirow{7}{*}{\rotatebox{90}{Weather}}
&24&\textbf{0.098}&\textbf{0.131}&\underline{0.100}&0.135&0.102&0.137&0.107&\underline{0.133}&0.131&0.169&0.128&0.158&0.102&0.156&0.120&0.177&0.205&0.290&0.192&0.275\\
&48&\textbf{0.129}&\underline{0.172}&\underline{0.130}&0.176&0.131&0.177&\underline{0.130}&\textbf{0.161}&0.162&0.196&0.153&0.190&0.133&0.190&0.173&0.236&0.213&0.303&0.214&0.290\\
&96 &\textbf{0.160}&\underline{0.203}&\underline{0.162}&\textbf{0.197}&0.163&0.206&0.164&0.210&0.180&0.218&0.172&0.215&0.171&0.226&0.203&0.268&0.266&0.341&0.228&0.297\\
&192&\textbf{0.205}&\textbf{0.250}&0.211&0.253&0.210&0.253&\underline{0.209}&\underline{0.251}&0.223&0.257&0.216&0.253&0.213&0.279&0.237&0.294&0.275&0.329&0.280&0.339\\
&336&\textbf{0.262}&\textbf{0.291}&0.269&0.295&0.269&0.295&\underline{0.265}&\underline{0.294}&0.283&0.299&0.275&0.295&0.277&0.341&0.291&0.345&0.342&0.371&0.326&0.367\\
&720&\textbf{0.341}&\textbf{0.342}&0.352&0.347&0.347&0.347&\underline{0.344}&\underline{0.345}&0.358&0.349&0.358&0.349&0.354&0.382&0.355&0.395&0.386&0.414&0.399&0.409\\
\cmidrule(lr){2-22}
&Avg&\textbf{0.199}&\textbf{0.231}&0.204&0.233&\underline{0.203}&0.235&\underline{0.203}&\underline{0.232}&0.222&0.248&0.217&0.243&0.207&0.262&0.229&0.285&0.281&0.341&0.273&0.329\\

\midrule[0.7pt]

\multirow{7}{*}{\rotatebox{90}{Electricity}}
&24&\textbf{0.115}&\textbf{0.214}&\underline{0.118}&\underline{0.216}&0.119&0.219&0.128&0.224&0.126&0.231&0.138&0.228&0.126&0.231&0.169&0.254&0.229&0.343&0.240&0.341\\
&48&\textbf{0.140}&\textbf{0.242}&0.147&0.245&\underline{0.144}&\underline{0.244}&0.148&0.245&0.153&0.254&0.161&0.261&0.154&0.259&0.198&0.274&0.275&0.373&0.259&0.368\\
&96&\textbf{0.155}&\textbf{0.247}&0.164&0.252&0.160&0.257&0.162&0.256&\underline{0.159}&\underline{0.250}&0.177&0.266&0.165&0.268&0.194&0.276&0.283&0.381&0.273&0.382\\
&192&\textbf{0.169}&\textbf{0.251}&0.180&0.266&\underline{0.175}&0.270&\underline{0.175}&\underline{0.265}&0.182&0.312&0.188&0.280&0.177&0.278&0.195&0.281&0.301&0.397&0.302&0.397\\
&336&\textbf{0.185}&\textbf{0.277}&0.193&\textbf{0.277}&0.197&0.293&\underline{0.190}&\underline{0.281}&0.200&0.287&0.200&0.290&0.192&0.293&0.207&0.295&0.286&0.384&0.310&0.402\\
&720&\textbf{0.221}&\textbf{0.305}&\underline{0.224}&\underline{0.309}&0.249&0.333&0.229&0.317&0.226&0.312&0.241&0.325&0.263&0.339&0.242&0.328&0.329&0.409&0.377&0.449\\
\cmidrule(lr){2-22}
&Avg&\textbf{0.164}&\textbf{0.256}&\underline{0.171}&\underline{0.260}&0.174&0.269&0.172&0.264&0.174&0.274&0.184&0.275&0.179&0.278&0.200&0.284&0.283&0.381&0.293&0.389\\

\bottomrule[1.pt]
\end{tabular}
}
\caption{Performance comparison with baselines. \textbf{Bold} \& \underline{Underline} indicate the best \& second best results, respectively.}
\label{result}
\end{table}

\subsection{Result Analysis}

Table \ref{result} presents the comparison results, from which we can observe that our method exhibits substantial improvements across a majority of cases in comparison to both MLP-based and Transformer-based methods. Specifically, across the seven datasets, the average MSE over six prediction windows decreased by 3.84\%, 7.45\%, 6.41\%, 1.55\%, 1.52\%, 1.97\%, and 4.09\% compared to the best-performing baselines. In addition, the MAE was also reduced by 1.89\%, 6.84\%, 1.43\%, 0.83\%, 1.88\%, 0.43\% and 1.53\%, respectively. Moreover, TimeMixer++, PatchMLP, TimeMixer, iTransformer, and PatchTST achieve second-best average results across different datasets, primarily because they still struggle to capture semantic dependencies related to temporal characteristics. These advancements highlight the importance of TimeFormer in enhancing semantic representations and modeling temporal dependencies in time series data, thereby enabling superior performance compared to state-of-the-art methods.

\subsection{Ablation Study}
We conduct ablation studies to evaluate the contributions of the semantic enhancement module and the proposed self-attention mechanism (MoSA) within TimeFormer, corresponding to two variants. In the first variant ``w/o SS'', we exclude the usage of subsequence segmentation. Moreover, to verify the effectiveness of the proposed MoSA, the second variant ``w/o MoSA'', replaces MoSA with the standard self-attention mechanism of the Transformer. Finally, we compare them with the complete TimeFormer.

Table \ref{Ablation} presents the results of the ablation study conducted on two datasets (ETTm1 and ETTh1) across six different prediction lengths. We can observe that the complete TimeFormer outperforms its variants, which indicates that each component contributes positively to the prediction. Specifically, compared to the variant ``w/o SS'', our complete TimeFormer achieves average drops of 14.24\% and 7.51\% in terms of MSE on the two datasets. This suggests that merely modeling temporal dependencies across the entire time series may overlook the semantic information within the time series. Moreover, we see decreases of 3.56\% and 4.13\% in terms of MSE by applying MoSA. This can be attributed to the combined introduction of the causal mask and the Hawkes process, which simulate the temporal influence. Overall, the results underscore the necessity of the subsequence analysis approach and MoSA, both of which are tailored to the unique characteristics of time series data.

\begin{table}[t]

\renewcommand{\arraystretch}{1}
\footnotesize	
\centering
\resizebox{0.7\linewidth}{!}{
\begin{tabular}
{c|c|cc|cc|cc}
\toprule[1.pt]

\multicolumn{2}{c}{Methods}& \multicolumn{2}{c}{w/o SS}& \multicolumn{2}{c}{w/o MoSA}&\multicolumn{2}{c}{TimeFormer}\\ 
\cmidrule(lr){1-2}\cmidrule(lr){3-4}\cmidrule(lr){5-6}\cmidrule(lr){7-8}
\multicolumn{2}{c}{Metric}& MSE&MAE& MSE&MAE& MSE&MAE\\
\toprule[1.pt]
\multirow{7}{*}{\rotatebox{90}{ETTm1}}
& 24&0.243&0.315&0.214&0.291&\textbf{0.204}&\textbf{0.281}\\
& 48&0.315&0.348&0.291&0.343&\textbf{0.275}&\textbf{0.329}\\
& 96&0.373&0.392&0.323&0.368&\textbf{0.311}&\textbf{0.354}\\
&192&0.396&0.404&0.353&0.390&\textbf{0.349}&\textbf{0.377}\\
&336&0.449&0.437&0.396&0.421&\textbf{0.377}&\textbf{0.399}\\
&720&0.502&0.489&0.447&0.445&\textbf{0.439}&\textbf{0.436}\\
\cmidrule(lr){2-8}
&Avg&0.379&0.397&0.337&0.378&\textbf{0.325}&\textbf{0.362}\\

\midrule[0.7pt]

\multirow{7}{*}{\rotatebox{90}{ETTh1}}
& 24&0.333&0.382&0.314&0.357&\textbf{0.292}&\textbf{0.349}\\
& 48&0.374&0.405&0.356&0.382&\textbf{0.335}&\textbf{0.374}\\
& 96&0.402&0.434&0.382&0.414&\textbf{0.374}&\textbf{0.398}\\
&192&0.450&0.462&0.454&0.439&\textbf{0.431}&\textbf{0.434}\\
&336&0.495&0.487&0.475&0.463&\textbf{0.468}&\textbf{0.457}\\
&720&0.507&0.492&0.488&0.481&\textbf{0.464}&\textbf{0.470}\\
\cmidrule(lr){2-8}
&Avg&0.426&0.443&0.411&0.422&\textbf{0.394}&\textbf{0.413}\\

\bottomrule[1.pt]
\end{tabular}
}
\caption{Ablation study.}
\label{Ablation}
\end{table}

\subsection{Analysis of Self-attention with Modulation Term}
We further examine MoSA from three distinct perspectives to assess its ability to capture temporal dependencies: visualization analysis, performance comparison with self-attention (SA) and complexity analysis.

\begin{figure}[t]
    \centering
    \begin{minipage}{0.45\textwidth}
        \centering
        \includegraphics[width=\textwidth]{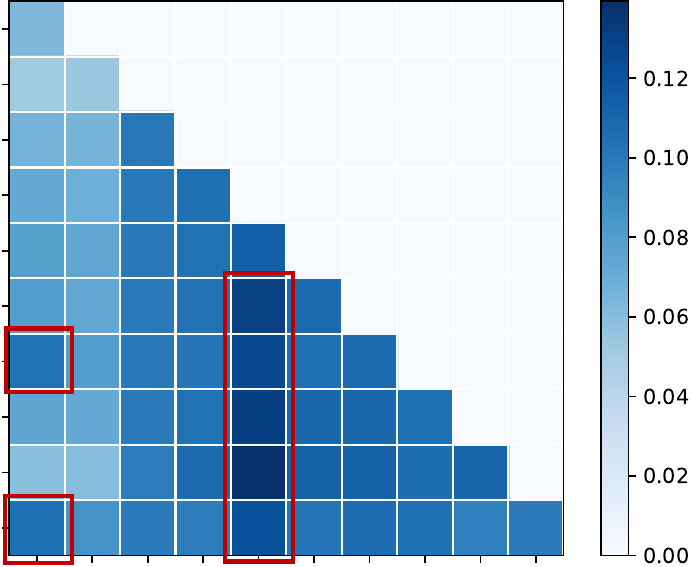}
        \subcaption{Intra-patch correlations of MoSA.}\label{figsub1}
    \end{minipage}\hspace{0.005\textwidth}
    \begin{minipage}{0.45\textwidth}
        \centering
        \includegraphics[width=\textwidth]{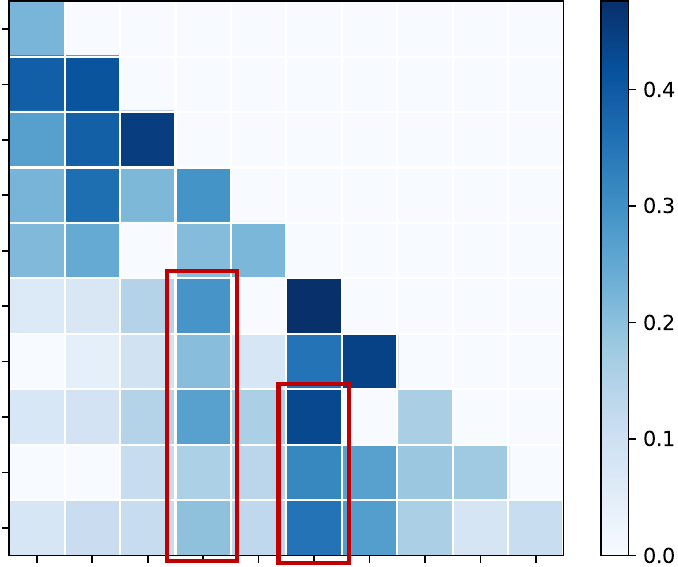}
        \subcaption{Inter-patch correlations of MoSA.}\label{figsub2}
    \end{minipage}
    \caption{Visualization analysis of temporal attention matrices.}
    \label{visilization}
\end{figure}

\subsubsection{Visualization Analysis}
We conduct experiments on the ETTh1 dataset and visualize the intra- and inter-sequence temporal attention matrices generated from TimeFormer, with $S=1$. Figure~\ref{visilization} illustrates the visualization results. Specifically, in the matrix of Figure~\ref{figsub1}, the columns on the left represent earlier time steps of a subsequence, while in the matrix of Figure~\ref{figsub2}, the columns on the left represent earlier subsequences. First, we can observe that all non-zero elements in the two matrices are confined to the lower triangular part, which suggests that during the feature aggregation process (i.e., Eq.\eqref{aggregation}), the features at the current time are solely influenced by previous time steps, thereby achieving temporal causality. Moreover, it can be observed from both matrices that the values of elements on the left are generally lower than those on the right. This indicates the decay effect that temporal influence decline over time. We can also observe that there are still several larger values at the leftmost time steps (see the red boxes), which suggests that while TimeFormer captures temporal dependencies, it also preserves the ability to extract global pairwise dependencies between tokens.

\subsubsection{Performance Comparison with Self-attention}

\begin{table}[t]

\renewcommand{\arraystretch}{1}
\footnotesize	
\centering
\setlength{\tabcolsep}{3pt}
\resizebox{0.7\linewidth}{!}{ 
\begin{tabular}
{c|c|cc|cc|cc|cc}
\toprule[1.pt]

\multicolumn{2}{c}{Methods}& \multicolumn{2}{c}{PatchTST}& \multicolumn{2}{c}{+MoSA}&\multicolumn{2}{c}{Transformer}& \multicolumn{2}{c}{+MoSA}\\ 
\cmidrule(lr){1-2}\cmidrule(lr){3-4}\cmidrule(lr){5-6}\cmidrule(lr){7-8}\cmidrule(lr){9-10}
\multicolumn{2}{c}{Metric}& MSE&MAE& MSE&MAE& MSE&MAE& MSE&MAE\\
\toprule[1.pt]
\multirow{7}{*}{\rotatebox{90}{ETTm1}}
&24&0.214&0.288&\textbf{0.210}&\textbf{0.285}&0.291&0.362&\textbf{0.274}&\textbf{0.346}\\
&48&0.293&0.341&\textbf{0.285}&\textbf{0.338}&0.528&0.507&\textbf{0.397}&\textbf{0.432}\\
&96&0.329&0.367&\textbf{0.322}&\textbf{0.362}&0.590&0.556&\textbf{0.486}&\textbf{0.493}\\
&192&0.367&0.387&\textbf{0.360}&\textbf{0.374}&0.786&0.647&\textbf{0.700}&\textbf{0.617}\\
&336&0.393&0.406&\textbf{0.385}&\textbf{0.401}&0.968&0.773&\textbf{0.938}&\textbf{0.767}\\
&720&0.453&0.441&\textbf{0.446}&\textbf{0.447}&1.162&0.818&\textbf{0.957}&\textbf{0.753}\\
\cmidrule(lr){2-10}
&Avg&0.341&0.371&\textbf{0.334}&\textbf{0.367}&0.721&0.610&\textbf{0.625}&\textbf{0.568}\\

\midrule[0.7pt]

\multirow{7}{*}{\rotatebox{90}{ETTh1}}
&24&0.307&0.359&\textbf{0.299}&\textbf{0.352}&0.585&0.558&\textbf{0.517}&\textbf{0.518}\\
&48&0.353&0.388&\textbf{0.343}&\textbf{0.379}&0.732&0.647&\textbf{0.600}&\textbf{0.581}\\
&96&0.389&0.409&\textbf{0.383}&\textbf{0.400}&0.771&0.686&\textbf{0.732}&\textbf{0.653}\\
&192&0.442&0.440&\textbf{0.436}&\textbf{0.435}&0.872&0.755&\textbf{0.823}&\textbf{0.706}\\
&336&0.495&0.469&\textbf{0.483}&\textbf{0.461}&0.946&0.780&\textbf{0.889}&\textbf{0.747}\\
&720&0.502&0.489&\textbf{0.487}&\textbf{0.466}&1.113&0.867&\textbf{1.020}&\textbf{0.813}\\
\cmidrule(lr){2-10}
&Avg&0.414&0.425&\textbf{0.405}&\textbf{0.415}&0.836&0.715&\textbf{0.763}&\textbf{0.669}\\

\bottomrule[1.pt]
\end{tabular}}
\caption{Evaluation of the applicability of MoSA.}
\label{BiaSAAnalysis}
\end{table}

To further evaluate the effectiveness of MoSA, we replace the conventional self-attention (SA) in Transformer-based models with MoSA, and compare their performance. Specifically, we select PatchTST and the vanilla Transformer as representative models for comparison. Specifically, in PatchTST, each patch is regarded as an individual token, whereas in the Transformer, each time step is treated as a token. It is noted that, due to the absence of temporal dependencies between variables, MoSA is not applicable to iTransformer. Table \ref{BiaSAAnalysis} illustrates the comparison results on ETTm1 and ETTh1 datasets. Specifically, for PatchTST, employing MoSA results in superior performance compared to SA across all prediction lengths. On the two datasets, the average MSE decreased by 2.05\% and 2.17\%, respectively, while the average MAE decreased by 1.07\% and 2.35\%, respectively. Similar observation can be obtained in the vanilla Transformer. Specifically, MoSA reduces the average MSE by 13.31\% and 8.73\% and the average MAE by 6.88\% and 6.43\% on the two datasets, respectively. This further confirms MoSA’s effectiveness in capturing temporal dependencies, demonstrating its potential as a plug-in to enhance Transformer-based time series predictors with strong applicability.

\subsubsection{Complexity Analysis}

The computational complexity of the self-attention is driven by the computation of the attention matrix and the weighted sum of the values, i.e., $\mathcal{O}(T^2\times d)$, where $T$ is the number of tokens and $d$ is the dimension. Compared to self-attention, the proposed MoSA introduces an additional temporal causality mask and a Hawkes process-inspired weighting. Thus, the total computational complexity consumes $\mathcal{O}(T^2\times d)+\mathcal{O}(T^2)+\mathcal{O}(T^2)$. The increase of computational complexity lies in the first term $\mathcal{O}(T^2\times d)$. The additional computational cost is acceptable.

\begin{figure}[t]
    \centering
    \begin{minipage}{0.45\textwidth}
        \centering
        \includegraphics[width=\textwidth]{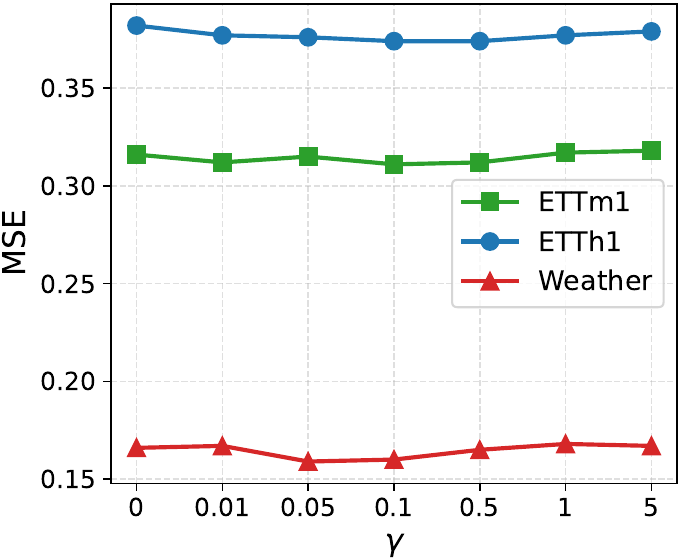}
        \subcaption{Sensitivity analysis for $\gamma$.}\label{fig:sub1}
    \end{minipage}\hspace{0.005\textwidth}
    \begin{minipage}{0.45\textwidth}
        \centering
        \includegraphics[width=\textwidth]{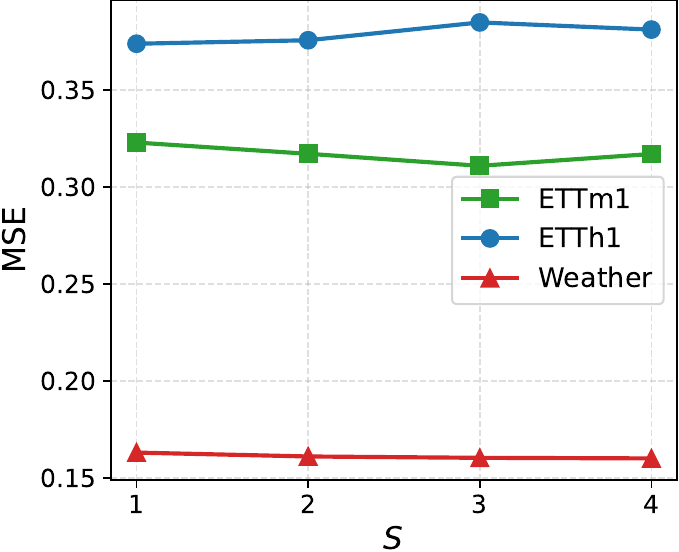}
        \subcaption{Sensitivity analysis for $S$.}\label{fig:sub2}
    \end{minipage}
    \caption{Sensitivity analysis.}
    \label{hyper}
\end{figure}

\subsection{Hyperparameter Sensitivity Analysis}

\subsubsection{Sensitivity analysis for $\gamma$}
We utilize the Hawkes process to enhance the attention matrix by explicitly modeling the temporal decay effect in time series. Given the critical role of the decay rate $\gamma$, we explore $\gamma$ values within the range of $\{0, 0.01, 0.05, 0.1, 0.5, 1, 5\}$. Specifically, $\gamma = 0$ represents to the case without modeling the decay effect. As shown in Figure \ref{fig:sub1}, the optimal performance is achieved with $\gamma = 0.05$ for the Weather dataset and $\gamma = 0.1$ for the ETTh1 and ETTm1 datasets. When $\gamma$ is too small (e.g., 0 or 0.01), the decay effect becomes negligible and is not effectively captured. Conversely, when $\gamma$ is too large, the attention weights assigned to earlier time steps become excessively diminished, leading to the underutilization of historical information.

\subsubsection{Sensitivity analysis for $S$}
We perform a sensitivity analysis for the number of scales $S$. As shown in Figure \ref{fig:sub2}, the optimal value of $S$ varies across different datasets. Specifically, for the ETTh1 dataset, the best performance is achieved when $S=1$, while for the ETTm1 dataset, $S=3$ yields the best results. On the Weather dataset, $S=4$ provides the optimal performance. For the minute-level ETTm1 dataset, additional downsampling enables the consideration of richer coarse grained information (e.g., hourly patterns). Moreover, compared to the ETT datasets, minute-level weather data exhibits more hourly and daily variations, thereby requiring more sampling layers to capture temporal patterns.

\section{Conclusion}
\label{conclusion}

This paper presents TimeFormer, a model specifically designed for time series forecasting. The most innovative aspect of TimeFormer is its Modulated Self-Attention (MoSA), which incorporates modulation terms based on the Hawkes process and causal masking strategies to constrain the learning process of attention coefficients. MoSA features two characteristics and a coupling of temporal priors: unidirectional attention from past to future and weight decay, driven by the inherent priors of time series data. Additionally, we propose a framework based on multi-scale and subsequence analysis that utilizes MoSA to enhance semantic representations within subsequences and reveal semantic relationships between subsequences at different temporal scales, thereby capturing rich temporal dependencies. We conduct extensive experiments to evaluate our method, demonstrating its effectiveness in modeling temporal dependencies within time series data. Furthermore, MoSA serves as a generic plug-in that can be incorporated into various Transformer-based models to improve prediction accuracy.

\bibliographystyle{cas-model2-names}

\bibliography{cas-refs}

\begin{thebibliography}{42}
\expandafter\ifx\csname natexlab\endcsname\relax\def\natexlab#1{#1}\fi
\providecommand{\url}[1]{\texttt{#1}}
\providecommand{\href}[2]{#2}
\providecommand{\path}[1]{#1}
\providecommand{\DOIprefix}{doi:}
\providecommand{\ArXivprefix}{arXiv:}
\providecommand{\URLprefix}{URL: }
\providecommand{\Pubmedprefix}{pmid:}
\providecommand{\doi}[1]{\href{http://dx.doi.org/#1}{\path{#1}}}
\providecommand{\Pubmed}[1]{\href{pmid:#1}{\path{#1}}}
\providecommand{\bibinfo}[2]{#2}
\ifx\xfnm\relax \def\xfnm[#1]{\unskip,\space#1}\fi
\bibitem[{Adelaida Ojeda~Beltran(2025)}]{Beltran2025series}
\bibinfo{author}{Adelaida Ojeda~Beltran, Mario E. Suaza-Medina,
  F.J.Z.S.E.D.L.H.F.J.E.G.}, \bibinfo{year}{2025}.
\newblock \bibinfo{title}{Multivariate integration of time series with ml for
  corn price forecasting in colombia}.
\newblock \bibinfo{journal}{Expert Systems with Applications} ,
  \bibinfo{pages}{129822}.
\bibitem[{Chaoyang~Wang(2025)}]{Wang2025TimeParticle}
\bibinfo{author}{Chaoyang~Wang, G.L.}, \bibinfo{year}{2025}.
\newblock \bibinfo{title}{Timeparticle: Particle-like multiscale state space
  models for time series forecasting}.
\newblock \bibinfo{journal}{Knowledge-Based Systems} \bibinfo{volume}{325},
  \bibinfo{pages}{113923}.
\bibitem[{Chen et~al.(2024)Chen, Zhang, Cheng, Shu, Wang, Wen, Yang and
  Guo}]{chen2024pathformer}
\bibinfo{author}{Chen, P.}, \bibinfo{author}{Zhang, Y.},
  \bibinfo{author}{Cheng, Y.}, \bibinfo{author}{Shu, Y.},
  \bibinfo{author}{Wang, Y.}, \bibinfo{author}{Wen, Q.}, \bibinfo{author}{Yang,
  B.}, \bibinfo{author}{Guo, C.}, \bibinfo{year}{2024}.
\newblock \bibinfo{title}{Pathformer: Multi-scale transformers with adaptive
  pathways for time series forecasting}.
\newblock \bibinfo{journal}{arXiv preprint arXiv:2402.05956} .
\bibitem[{Chen et~al.(2023)Chen, Ma, Li, Wang and Li}]{chen2023long}
\bibinfo{author}{Chen, Z.}, \bibinfo{author}{Ma, M.}, \bibinfo{author}{Li, T.},
  \bibinfo{author}{Wang, H.}, \bibinfo{author}{Li, C.}, \bibinfo{year}{2023}.
\newblock \bibinfo{title}{Long sequence time-series forecasting with deep
  learning: A survey}.
\newblock \bibinfo{journal}{Information Fusion} \bibinfo{volume}{97},
  \bibinfo{pages}{101819}.
\bibitem[{De~Livera et~al.(2011)De~Livera, Hyndman and
  Snyder}]{de2011forecasting}
\bibinfo{author}{De~Livera, A.M.}, \bibinfo{author}{Hyndman, R.J.},
  \bibinfo{author}{Snyder, R.D.}, \bibinfo{year}{2011}.
\newblock \bibinfo{title}{Forecasting time series with complex seasonal
  patterns using exponential smoothing}.
\newblock \bibinfo{journal}{Journal of the American statistical association}
  \bibinfo{volume}{106}, \bibinfo{pages}{1513--1527}.
\bibitem[{Denghui~Xu(2025)}]{Xu2025FPF}
\bibinfo{author}{Denghui~Xu, Hua~Wang, F.Z.}, \bibinfo{year}{2025}.
\newblock \bibinfo{title}{Frequency decomposition and patch modeling framework
  for time-series forecasting}.
\newblock \bibinfo{journal}{Applied Soft Computing} \bibinfo{volume}{185},
  \bibinfo{pages}{113890}.
\bibitem[{Dunlu~Peng(2026)}]{Peng2026ConEm}
\bibinfo{author}{Dunlu~Peng, Q.L.}, \bibinfo{year}{2026}.
\newblock \bibinfo{title}{Samforecast: A hybrid model of self-attention and
  mamba with adaptive wavelet transform for time series forecasting}.
\newblock \bibinfo{journal}{Expert Systems with Applications}
  \bibinfo{volume}{298}, \bibinfo{pages}{129498}.
\bibitem[{Gang~Tan(2025a)}]{Tan2025series}
\bibinfo{author}{Gang~Tan, Yueyang~Wang, Z.X.D.H.G.S.}, \bibinfo{year}{2025}a.
\newblock \bibinfo{title}{From a multi-period perspective: A periodic dynamics
  forecasting network for multivariate time series forecasting}.
\newblock \bibinfo{journal}{Pattern Recognition} \bibinfo{volume}{167},
  \bibinfo{pages}{111760}.
\bibitem[{Gang~Tan(2025b)}]{Luo2025TFDNet}
\bibinfo{author}{Gang~Tan, Yueyang~Wang, Z.X.D.H.G.S.}, \bibinfo{year}{2025}b.
\newblock \bibinfo{title}{Tfdnet: Time–frequency enhanced decomposed network
  for long-term time series forecasting}.
\newblock \bibinfo{journal}{Pattern Recognition} \bibinfo{volume}{162},
  \bibinfo{pages}{111412}.
\bibitem[{Guangbao~Zhou(2026)}]{Zhou2026TAMD}
\bibinfo{author}{Guangbao~Zhou, Pengliang~Liu, Q.L.M.Q.Z.X.Z.Z.L.L.},
  \bibinfo{year}{2026}.
\newblock \bibinfo{title}{Time series adaptive mode decomposition (tamd):
  Method for improving forecasting accuracy in the apparel industry}.
\newblock \bibinfo{journal}{Pattern Recognition} \bibinfo{volume}{172},
  \bibinfo{pages}{112417}.
\bibitem[{Hoang Nguyen~Nguyen(2025)}]{Nguyen2025ConEm}
\bibinfo{author}{Hoang Nguyen~Nguyen, Wei~Xiang, L.C.M.D.G.S.A.M.T.L.Y.},
  \bibinfo{year}{2025}.
\newblock \bibinfo{title}{Conem: A novel framework for integrating external
  factors with inner and outer correlations in time series forecasting}.
\newblock \bibinfo{journal}{Knowledge-Based Systems} \bibinfo{volume}{329},
  \bibinfo{pages}{114312}.
\bibitem[{Huang et~al.(2024)Huang, Shen, Zhang, Cheng, Ding, Zhou and
  Wang}]{huang2024hdmixer}
\bibinfo{author}{Huang, Q.}, \bibinfo{author}{Shen, L.},
  \bibinfo{author}{Zhang, R.}, \bibinfo{author}{Cheng, J.},
  \bibinfo{author}{Ding, S.}, \bibinfo{author}{Zhou, Z.},
  \bibinfo{author}{Wang, Y.}, \bibinfo{year}{2024}.
\newblock \bibinfo{title}{Hdmixer: Hierarchical dependency with extendable
  patch for multivariate time series forecasting}, in:
  \bibinfo{booktitle}{Proceedings of the AAAI conference on artificial
  intelligence}, pp. \bibinfo{pages}{12608--12616}.
\bibitem[{Huang et~al.(2023)Huang, Shen, Zhang, Ding, Wang, Zhou and
  Wang}]{huang2023crossgnn}
\bibinfo{author}{Huang, Q.}, \bibinfo{author}{Shen, L.},
  \bibinfo{author}{Zhang, R.}, \bibinfo{author}{Ding, S.},
  \bibinfo{author}{Wang, B.}, \bibinfo{author}{Zhou, Z.},
  \bibinfo{author}{Wang, Y.}, \bibinfo{year}{2023}.
\newblock \bibinfo{title}{Crossgnn: Confronting noisy multivariate time series
  via cross interaction refinement}.
\newblock \bibinfo{journal}{Advances in Neural Information Processing Systems}
  \bibinfo{volume}{36}, \bibinfo{pages}{46885--46902}.
\bibitem[{Junjie~Ye(2026)}]{Ye2026CVACL-MA}
\bibinfo{author}{Junjie~Ye, Chengli~Zhou, X.Z.Y.H.C.Z.}, \bibinfo{year}{2026}.
\newblock \bibinfo{title}{Cvacl-ma: Comprehensive variate analysis and
  collaborative learning with multi-adapter for multivariate time series
  forecasting}.
\newblock \bibinfo{journal}{Pattern Recognition} \bibinfo{volume}{169},
  \bibinfo{pages}{111936}.
\bibitem[{Kalpakis et~al.(2001)Kalpakis, Gada and
  Puttagunta}]{kalpakis2001distance}
\bibinfo{author}{Kalpakis, K.}, \bibinfo{author}{Gada, D.},
  \bibinfo{author}{Puttagunta, V.}, \bibinfo{year}{2001}.
\newblock \bibinfo{title}{Distance measures for effective clustering of arima
  time-series}, in: \bibinfo{booktitle}{Proceedings 2001 IEEE international
  conference on data mining}, \bibinfo{organization}{IEEE}. pp.
  \bibinfo{pages}{273--280}.
\bibitem[{Kingma and Ba(2014)}]{kingma2014adam}
\bibinfo{author}{Kingma, D.P.}, \bibinfo{author}{Ba, J.}, \bibinfo{year}{2014}.
\newblock \bibinfo{title}{Adam: A method for stochastic optimization}.
\newblock \bibinfo{journal}{arXiv preprint arXiv:1412.6980} .
\bibitem[{Kong et~al.(2025)Kong, Wang, Nie, Zhou, Zohren, Liang, Sun and
  Wen}]{kong2025unlocking}
\bibinfo{author}{Kong, Y.}, \bibinfo{author}{Wang, Z.}, \bibinfo{author}{Nie,
  Y.}, \bibinfo{author}{Zhou, T.}, \bibinfo{author}{Zohren, S.},
  \bibinfo{author}{Liang, Y.}, \bibinfo{author}{Sun, P.}, \bibinfo{author}{Wen,
  Q.}, \bibinfo{year}{2025}.
\newblock \bibinfo{title}{Unlocking the power of lstm for long term time series
  forecasting}, in: \bibinfo{booktitle}{Proceedings of the AAAI Conference on
  Artificial Intelligence}, pp. \bibinfo{pages}{11968--11976}.
\bibitem[{Lim and Zohren(2021)}]{lim2021time}
\bibinfo{author}{Lim, B.}, \bibinfo{author}{Zohren, S.}, \bibinfo{year}{2021}.
\newblock \bibinfo{title}{Time-series forecasting with deep learning: a
  survey}.
\newblock \bibinfo{journal}{Philosophical Transactions of the Royal Society A}
  \bibinfo{volume}{379}, \bibinfo{pages}{20200209}.
\bibitem[{Liu et~al.(2023)Liu, Hu, Zhang, Wu, Wang, Ma and
  Long}]{liu2023itransformer}
\bibinfo{author}{Liu, Y.}, \bibinfo{author}{Hu, T.}, \bibinfo{author}{Zhang,
  H.}, \bibinfo{author}{Wu, H.}, \bibinfo{author}{Wang, S.},
  \bibinfo{author}{Ma, L.}, \bibinfo{author}{Long, M.}, \bibinfo{year}{2023}.
\newblock \bibinfo{title}{itransformer: Inverted transformers are effective for
  time series forecasting}.
\newblock \bibinfo{journal}{arXiv preprint arXiv:2310.06625} .
\bibitem[{Liu et~al.(2025a)Liu, Duan, Chu, Kuhlmann, Zhang, Yue, Tang and
  Zhang}]{liu2025attributed}
\bibinfo{author}{Liu, Z.}, \bibinfo{author}{Duan, P.}, \bibinfo{author}{Chu,
  Q.}, \bibinfo{author}{Kuhlmann, L.}, \bibinfo{author}{Zhang, C.},
  \bibinfo{author}{Yue, W.}, \bibinfo{author}{Tang, X.},
  \bibinfo{author}{Zhang, B.}, \bibinfo{year}{2025}a.
\newblock \bibinfo{title}{An attributed multiplex network enabled gnn-based
  stock predictor with observable and non-observable information}.
\newblock \bibinfo{journal}{Expert Systems with Applications} ,
  \bibinfo{pages}{129018}.
\bibitem[{Liu et~al.(2025b)Liu, Duan, Wang, Tang, Chu, Zhang, Huang and
  Zhang}]{liu2025disms}
\bibinfo{author}{Liu, Z.}, \bibinfo{author}{Duan, P.}, \bibinfo{author}{Wang,
  B.}, \bibinfo{author}{Tang, X.}, \bibinfo{author}{Chu, Q.},
  \bibinfo{author}{Zhang, C.}, \bibinfo{author}{Huang, Y.},
  \bibinfo{author}{Zhang, B.}, \bibinfo{year}{2025}b.
\newblock \bibinfo{title}{Disms-ts: Eliminating redundant multi-scale features
  for time series classification}.
\newblock \bibinfo{journal}{arXiv preprint arXiv:2507.04600} .
\bibitem[{Nie et~al.(2022)Nie, Nguyen, Sinthong and Kalagnanam}]{nie2022time}
\bibinfo{author}{Nie, Y.}, \bibinfo{author}{Nguyen, N.H.},
  \bibinfo{author}{Sinthong, P.}, \bibinfo{author}{Kalagnanam, J.},
  \bibinfo{year}{2022}.
\newblock \bibinfo{title}{A time series is worth 64 words: Long-term
  forecasting with transformers}.
\newblock \bibinfo{journal}{arXiv preprint arXiv:2211.14730} .
\bibitem[{Qiu et~al.(2024)Qiu, Hu, Zhou, Wu, Du, Zhang, Guo, Zhou, Jensen,
  Sheng et~al.}]{qiu2024tfb}
\bibinfo{author}{Qiu, X.}, \bibinfo{author}{Hu, J.}, \bibinfo{author}{Zhou,
  L.}, \bibinfo{author}{Wu, X.}, \bibinfo{author}{Du, J.},
  \bibinfo{author}{Zhang, B.}, \bibinfo{author}{Guo, C.},
  \bibinfo{author}{Zhou, A.}, \bibinfo{author}{Jensen, C.S.},
  \bibinfo{author}{Sheng, Z.}, et~al., \bibinfo{year}{2024}.
\newblock \bibinfo{title}{Tfb: Towards comprehensive and fair benchmarking of
  time series forecasting methods}.
\newblock \bibinfo{journal}{arXiv preprint arXiv:2403.20150} .
\bibitem[{Shuai~Li(2025)}]{Li2025STNet}
\bibinfo{author}{Shuai~Li, Cheng~Zhao, J.Z.}, \bibinfo{year}{2025}.
\newblock \bibinfo{title}{Stnet: Seasonal-trend network for multivariate time
  series forecasting}.
\newblock \bibinfo{journal}{Neurocomputing} \bibinfo{volume}{655}.
\bibitem[{Tang and Zhang(2025)}]{tang2025unlocking}
\bibinfo{author}{Tang, P.}, \bibinfo{author}{Zhang, W.}, \bibinfo{year}{2025}.
\newblock \bibinfo{title}{Unlocking the power of patch: Patch-based mlp for
  long-term time series forecasting}, in: \bibinfo{booktitle}{Proceedings of
  the AAAI Conference on Artificial Intelligence}, pp.
  \bibinfo{pages}{12640--12648}.
\bibitem[{Tianxiang~Zhan(2024)}]{Zhan2024series}
\bibinfo{author}{Tianxiang~Zhan, F.X.}, \bibinfo{year}{2024}.
\newblock \bibinfo{title}{A novel weighted approach for time series forecasting
  based on visibility graph}.
\newblock \bibinfo{journal}{Pattern Recognition} \bibinfo{volume}{155},
  \bibinfo{pages}{110720}.
\bibitem[{Vaswani et~al.(2017)Vaswani, Shazeer, Parmar, Uszkoreit, Jones,
  Gomez, Kaiser and Polosukhin}]{vaswani2017attention}
\bibinfo{author}{Vaswani, A.}, \bibinfo{author}{Shazeer, N.},
  \bibinfo{author}{Parmar, N.}, \bibinfo{author}{Uszkoreit, J.},
  \bibinfo{author}{Jones, L.}, \bibinfo{author}{Gomez, A.N.},
  \bibinfo{author}{Kaiser, {\L}.}, \bibinfo{author}{Polosukhin, I.},
  \bibinfo{year}{2017}.
\newblock \bibinfo{title}{Attention is all you need}.
\newblock \bibinfo{journal}{Advances in neural information processing systems}
  \bibinfo{volume}{30}.
\bibitem[{Wang et~al.(2025)Wang, Mo, Xiang, Yin, Dai, Li and
  Fan}]{wang2025csformer}
\bibinfo{author}{Wang, H.}, \bibinfo{author}{Mo, Y.}, \bibinfo{author}{Xiang,
  K.}, \bibinfo{author}{Yin, N.}, \bibinfo{author}{Dai, H.},
  \bibinfo{author}{Li, B.}, \bibinfo{author}{Fan, S.}, \bibinfo{year}{2025}.
\newblock \bibinfo{title}{Csformer: Combining channel independence and mixing
  for robust multivariate time series forecasting}, in:
  \bibinfo{booktitle}{Proceedings of the AAAI Conference on Artificial
  Intelligence}, pp. \bibinfo{pages}{21090--21098}.
\bibitem[{Wang et~al.(2024a)Wang, Li, Shi, Ye, Mo, Lin, Ju, Chu and
  Jin}]{wang2024timemixer++}
\bibinfo{author}{Wang, S.}, \bibinfo{author}{Li, J.}, \bibinfo{author}{Shi,
  X.}, \bibinfo{author}{Ye, Z.}, \bibinfo{author}{Mo, B.},
  \bibinfo{author}{Lin, W.}, \bibinfo{author}{Ju, S.}, \bibinfo{author}{Chu,
  Z.}, \bibinfo{author}{Jin, M.}, \bibinfo{year}{2024}a.
\newblock \bibinfo{title}{Timemixer++: A general time series pattern machine
  for universal predictive analysis}.
\newblock \bibinfo{journal}{arXiv preprint arXiv:2410.16032} .
\bibitem[{Wang et~al.(2024b)Wang, Wu, Shi, Hu, Luo, Ma, Zhang and
  Zhou}]{wang2024timemixer}
\bibinfo{author}{Wang, S.}, \bibinfo{author}{Wu, H.}, \bibinfo{author}{Shi,
  X.}, \bibinfo{author}{Hu, T.}, \bibinfo{author}{Luo, H.},
  \bibinfo{author}{Ma, L.}, \bibinfo{author}{Zhang, J.Y.},
  \bibinfo{author}{Zhou, J.}, \bibinfo{year}{2024}b.
\newblock \bibinfo{title}{Timemixer: Decomposable multiscale mixing for time
  series forecasting}.
\newblock \bibinfo{journal}{arXiv preprint arXiv:2405.14616} .
\bibitem[{Wang et~al.(2024c)Wang, Feng, Qiu, Gu and Zhao}]{wang2024news}
\bibinfo{author}{Wang, X.}, \bibinfo{author}{Feng, M.}, \bibinfo{author}{Qiu,
  J.}, \bibinfo{author}{Gu, J.}, \bibinfo{author}{Zhao, J.},
  \bibinfo{year}{2024}c.
\newblock \bibinfo{title}{From news to forecast: Integrating event analysis in
  llm-based time series forecasting with reflection}.
\newblock \bibinfo{journal}{Advances in Neural Information Processing Systems}
  \bibinfo{volume}{37}, \bibinfo{pages}{58118--58153}.
\bibitem[{Wang et~al.(2024d)Wang, Wu, Dong, Qin, Zhang, Liu, Qiu, Wang and
  Long}]{wang2024timexer}
\bibinfo{author}{Wang, Y.}, \bibinfo{author}{Wu, H.}, \bibinfo{author}{Dong,
  J.}, \bibinfo{author}{Qin, G.}, \bibinfo{author}{Zhang, H.},
  \bibinfo{author}{Liu, Y.}, \bibinfo{author}{Qiu, Y.}, \bibinfo{author}{Wang,
  J.}, \bibinfo{author}{Long, M.}, \bibinfo{year}{2024}d.
\newblock \bibinfo{title}{Timexer: Empowering transformers for time series
  forecasting with exogenous variables}.
\newblock \bibinfo{journal}{arXiv preprint arXiv:2402.19072} .
\bibitem[{Wang et~al.(2024e)Wang, Xu, Yang, Wu, Li, Xie and
  Chen}]{wang2024fully}
\bibinfo{author}{Wang, Y.}, \bibinfo{author}{Xu, Y.}, \bibinfo{author}{Yang,
  J.}, \bibinfo{author}{Wu, M.}, \bibinfo{author}{Li, X.},
  \bibinfo{author}{Xie, L.}, \bibinfo{author}{Chen, Z.}, \bibinfo{year}{2024}e.
\newblock \bibinfo{title}{Fully-connected spatial-temporal graph for
  multivariate time-series data}, in: \bibinfo{booktitle}{Proceedings of the
  AAAI conference on artificial intelligence}, pp.
  \bibinfo{pages}{15715--15724}.
\bibitem[{Wang et~al.(2024f)Wang, Xu, Yang, Wu, Li, Xie and
  Chen}]{wang2024graph}
\bibinfo{author}{Wang, Y.}, \bibinfo{author}{Xu, Y.}, \bibinfo{author}{Yang,
  J.}, \bibinfo{author}{Wu, M.}, \bibinfo{author}{Li, X.},
  \bibinfo{author}{Xie, L.}, \bibinfo{author}{Chen, Z.}, \bibinfo{year}{2024}f.
\newblock \bibinfo{title}{Graph-aware contrasting for multivariate time-series
  classification}, in: \bibinfo{booktitle}{Proceedings of the AAAI conference
  on artificial intelligence}, pp. \bibinfo{pages}{15725--15734}.
\bibitem[{Wu et~al.(2022)Wu, Hu, Liu, Zhou, Wang and Long}]{wu2022timesnet}
\bibinfo{author}{Wu, H.}, \bibinfo{author}{Hu, T.}, \bibinfo{author}{Liu, Y.},
  \bibinfo{author}{Zhou, H.}, \bibinfo{author}{Wang, J.},
  \bibinfo{author}{Long, M.}, \bibinfo{year}{2022}.
\newblock \bibinfo{title}{Timesnet: Temporal 2d-variation modeling for general
  time series analysis}.
\newblock \bibinfo{journal}{arXiv preprint arXiv:2210.02186} .
\bibitem[{Wu et~al.(2021)Wu, Xu, Wang and Long}]{wu2021autoformer}
\bibinfo{author}{Wu, H.}, \bibinfo{author}{Xu, J.}, \bibinfo{author}{Wang, J.},
  \bibinfo{author}{Long, M.}, \bibinfo{year}{2021}.
\newblock \bibinfo{title}{Autoformer: Decomposition transformers with
  auto-correlation for long-term series forecasting}.
\newblock \bibinfo{journal}{Advances in neural information processing systems}
  \bibinfo{volume}{34}, \bibinfo{pages}{22419--22430}.
\bibitem[{Yang~Yu(2024)}]{Yu2024Robformer}
\bibinfo{author}{Yang~Yu, Ruizhe~Ma, Z.M.}, \bibinfo{year}{2024}.
\newblock \bibinfo{title}{Robformer: A robust decomposition transformer for
  long-term time series forecasting}.
\newblock \bibinfo{journal}{Pattern Recognition} \bibinfo{volume}{153},
  \bibinfo{pages}{110552}.
\bibitem[{Zeng et~al.(2023)Zeng, Chen, Zhang and Xu}]{zeng2023transformers}
\bibinfo{author}{Zeng, A.}, \bibinfo{author}{Chen, M.}, \bibinfo{author}{Zhang,
  L.}, \bibinfo{author}{Xu, Q.}, \bibinfo{year}{2023}.
\newblock \bibinfo{title}{Are transformers effective for time series
  forecasting?}, in: \bibinfo{booktitle}{Proceedings of the AAAI conference on
  artificial intelligence}, pp. \bibinfo{pages}{11121--11128}.
\bibitem[{Zhang and Yan(2023)}]{zhang2023crossformer}
\bibinfo{author}{Zhang, Y.}, \bibinfo{author}{Yan, J.}, \bibinfo{year}{2023}.
\newblock \bibinfo{title}{Crossformer: Transformer utilizing cross-dimension
  dependency for multivariate time series forecasting}, in:
  \bibinfo{booktitle}{The eleventh international conference on learning
  representations}.
\bibitem[{Zhong et~al.(2023)Zhong, Song, Zhuo, Li, Liu and
  Chan}]{zhong2023multi}
\bibinfo{author}{Zhong, S.}, \bibinfo{author}{Song, S.}, \bibinfo{author}{Zhuo,
  W.}, \bibinfo{author}{Li, G.}, \bibinfo{author}{Liu, Y.},
  \bibinfo{author}{Chan, S.H.G.}, \bibinfo{year}{2023}.
\newblock \bibinfo{title}{A multi-scale decomposition mlp-mixer for time series
  analysis}.
\newblock \bibinfo{journal}{arXiv preprint arXiv:2310.11959} .
\bibitem[{Zhou et~al.(2021)Zhou, Zhang, Peng, Zhang, Li, Xiong and
  Zhang}]{zhou2021informer}
\bibinfo{author}{Zhou, H.}, \bibinfo{author}{Zhang, S.}, \bibinfo{author}{Peng,
  J.}, \bibinfo{author}{Zhang, S.}, \bibinfo{author}{Li, J.},
  \bibinfo{author}{Xiong, H.}, \bibinfo{author}{Zhang, W.},
  \bibinfo{year}{2021}.
\newblock \bibinfo{title}{Informer: Beyond efficient transformer for long
  sequence time-series forecasting}, in: \bibinfo{booktitle}{Proceedings of the
  AAAI conference on artificial intelligence}, pp.
  \bibinfo{pages}{11106--11115}.
\bibitem[{Zhou et~al.(2022)Zhou, Ma, Wen, Wang, Sun and
  Jin}]{zhou2022fedformer}
\bibinfo{author}{Zhou, T.}, \bibinfo{author}{Ma, Z.}, \bibinfo{author}{Wen,
  Q.}, \bibinfo{author}{Wang, X.}, \bibinfo{author}{Sun, L.},
  \bibinfo{author}{Jin, R.}, \bibinfo{year}{2022}.
\newblock \bibinfo{title}{Fedformer: Frequency enhanced decomposed transformer
  for long-term series forecasting}, in: \bibinfo{booktitle}{International
  conference on machine learning}, \bibinfo{organization}{PMLR}. pp.
  \bibinfo{pages}{27268--27286}.

\end{thebibliography}

\end{document}